# Applying FrameNet to Chinese (Poetry)

Zirong Chen
*Department of Computer Science*
*Georgetown University*
Washington, DC, USA
zc157@gerogetown.edu

**Abstract**

FrameNet(Fillmore and Baker [2009]) is well-known for its wide use for knowledge representation in the form of inheritance-based ontologies and lexica(Trott et al. [2020]). Although FrameNet is usually applied to languages like English, Spanish and Italian, there are still plenty of FrameNet data sets available for other languages like Chinese, which differs significantly from those languages based on Latin alphabets. In this paper, the translation from ancient Chinese Poetry to modern Chinese will be first conducted to further apply the Chinese FrameNet(CFN, provided by Shanxi University). Afterwards, the translation from modern Chinese will be conducted as well for the comparison between the applications of CFN and English FrameNet. Finally, the overall comparison will be draw between CFN to modern Chinese and English FrameNet.

## 1. Introduction

The detailed definition of FrameNet will not be discussed in this paper.

CFN is a vocabulary database, including frames, vocabulary units and annotated sentences. It is based on the theory of frame semantics, cites Burke's work on FrameNet(Fillmore and Baker [2009]) written in English, and is supported by evidence from a large Chinese corpus. CFN currently contains 323 semantic frames, 3,947 lexical units, and more than 18,000 annotated sentences with framed syntax and semantic information, covering the common core of language and more specialized fields such as travel, online book sales and law. And a speech with 200 comments. In addition to constructing the CFN database, they also study the framework semantic theory related to Chinese, and study the construction technology of CFN-based applications. They have developed a frame semantic role labeling system for single sentences and speech.

Tangshi(Shi form of Chinese verse/Poetry in the Tang Dynasty, 618AD to 907AD), Songci(Ci form of Chinese verse/Poetry in the Song Dynasty, 960AD to 1279AD) and Yuanqu(Qu form of Chinese verse/Poetry in the Yuan Dynasty, 1271AD to 1368AD) are widely considered as the crowning achievement of Chinese Literature. Among them, Songci will be chosen for analysis in this paper, reasons are listed below.

Generally, there are two genres of Tangshi: seven-feet rhyme and five-feet rhyme. The seven-feet rhyme has 56 characters in total, with 8 sentences each containing 7 characters. Each character in Chinese has exactly 1 syllable, making each sentence 7-syllable long. Hence the name seven-feet rhyme. The five-feet rhyme works in a similar fashion. The appropriateness of grammar, especially by modern standards, was often sacrificed so that the meter and length of the poem could fit the strict fashion of Tangshi described above. So Tangshi is not suitable for the application of CFN. Because they were created as lyrics that are meant to fit into common tunes(there were more than 1,000 *types of tunes*), Songci and Yuanqu do not share the same constrains as Tangshi. Therefore, Songci and Yuanqu are more fitting options for CFN. Personally, I prefer Songci over Yuanqu due to historical reasons. Thus, I chose to apply Songci to CFN later.

## 2. Related Work

Previous work based on the FrameNet(Fillmore and Baker [2009]) project, such as the online annotation tool provided by the Brazil Research Lab, has made it more convenient than ever to do research on the different applications of FrameNet.

Besides, the Chinese FrameNet(CFN) is a current active research topic at Semantic Computing & Chinese FrameNet Research Center, Shanxi University, China. CFN currently contains 1,152 semantic frames, 12,153 lexical units(LU), more than 18,000 sentences. After consulting with Prof. Ru Li at Shanxi University, only the access to 1,152 frames and 12,153 lexical units is available for research proposes. With the help from CFN, we are now able to annotate modern Chinese translated from Songci.

## 3. Experiments

Experiments will include:

a) Translate Chinese Poetry to modern Chinese and English.

b) Apply FrameNet to annotate some sentences in modern Chinese and English.

c) Conclude some patterns in the application of FrameNet to Chinese.

d) Directly apply CFN to Chinese Poetry.

N.B. Chinese Poetry refers to ancient Chinese Poetry. More details are shown in Figure 1.

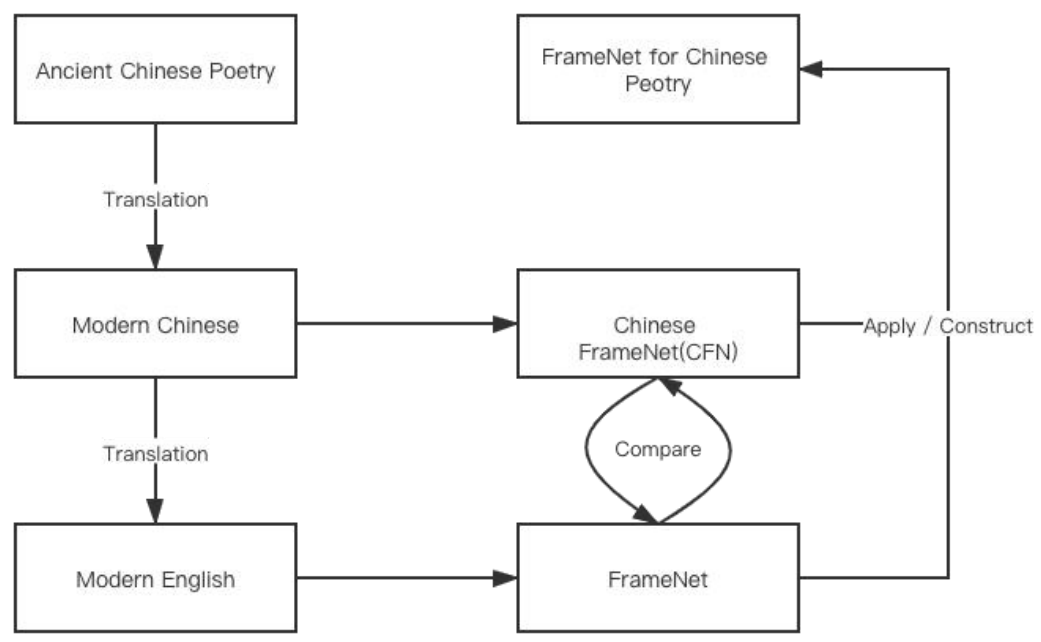

Figure 1

### 3.1 Experiment Resources

We will choose one piece of Songci in this paper to analyze and there are 9 sentences included in total:

雨霖铃·寒蝉凄切-柳永*(Yong Liu)*
*Yu/lin/lin · Han/chan/qi/qie*

N.B. We will apply '/' to separate the pinyin for each Chinese character.

Here is the poetry in Chinese('*Y·H*' for abbreviation):

雨霖铃·寒蝉凄切
[宋] 柳永

寒蝉凄切(a)，对长亭晚(b)，骤雨初歇(c)。[1]都门帐饮无绪(a)，留恋处，兰舟催发(b)。[2]执手相看泪眼，竟无语凝噎(a)。[3]念去去，千里烟波(a)，暮霭沉沉楚天阔(b)。[4]

多情自古伤离别(a)，更那堪，冷落清秋节(b)!
[5]今宵酒醒何处(a)？[6]杨柳岸，晓风残月(a)。[7]此去经年，应是良辰好景虚设(a)。[8]便纵有千种风情(a)，更与何人说(c)？[9]

### 3.2 Splitting and Translating

In this part, I set up two surveys for Chinese classmates：

1) is to split sentences in Chinese Poetry according to the context and its semantics.

2) is to translate sentences from Chinese Poetry to modern Chinese, and from modern Chinese to English.

The results of first survey is from Baidu Online Knowledge Base for Ancient Chinese Literature [2017]. And the final results of the second survey is based on the ranking of the peer reviewing scores, with a scale from 0 to 10, of all the answers from participants(three classmates, author and Google Translator). Ranking scores can be found in table 1, Appendix. Splitting results can be found in *Y·H*.

Here are the corresponding sub-sentences translated to modern Chinese:

[1]:
(a)秋蝉的叫声凄凉而急促
(b)傍晚时分，面对着长亭
(c)骤雨刚停

[2]:
(a)在京都郊外设帐饯行，却没有畅饮的心绪
(b)正在依依不舍的时候，船上的人已催着出发

[3]:
(a)握着对方的手含着泪对视，哽咽的说不出话来

[4]:
(a)想到这一去路途遥远，千里烟波渺茫
(b)傍晚的云雾笼罩着蓝天，深厚广阔，不知尽头

[5]:
(a)自古以来，多情的人总是为离别而伤感
(b)更何况是在这冷清、凄凉的秋天

[6]:
(a)谁知我今夜酒醒时身在何处

[7]:
(a)在杨柳岸边，伴着凄厉的晨风和黎明的残月

[8]:
(a)这一去长年相别，我料想即使遇到好天气、好风景，也如同虚设

[9]:
(a)即使有满腹的情意，又再同谁去诉说呢

Here are those sentences translated to English:

[1]:
(a)Autumn has fallen; the cicadas let out poignant and anxious cries.
(b)In the dusk, (I was) standing in front of the pavilion.

(c)The heavy rain suddenly stopped.

[2]:

(a)To bid you farewell, (I) had planned a dinner with you in a tent on the outskirts of the capital city, although (I) have neither the heart to drink nor eat.

(b)The boatman urged me to set off, although (we were) reluctant to say goodbye.

[3]:

(a)We held each others' hands, teary eyes staring, choked with sobs and speechless.

[4]:

(a)My mind is racing and I keep contemplating how the journey seems endless and how our fates seems uncertain.

(b)The nebulous mist of dusk covers the sky, hefty and wide, stretching beyond sight.

[5]:

(a)Throughout history, those who are most sensitive are also most distraught when departing one another.

(b)Not to mention that the departure happens to take place during fall, the season of melancholy.

[6]:

(a)Who knows where I would be tonight when I wake up from my drunken haze.

[7]:

(a)(Perhaps, I will wake up)At the bank under the willows, accompanied by the morning wind and crescent moon.

[8]:

(a)Our perennial separation renders any welcoming weather or beautiful scenery that I might encounter a mere illusion.

[9]:

(a)When I am full of affection in the future, with whom could I share my feelings?

### 3.3 Applying Chinese FrameNet

In order to be more demonstrative of the differences between the applications of FrameNet to different languages, only some of those sentences will be picked for further analysis: [1](b); [3](a); [8](a).

Before applying CFN to these sentences, the character-level tokenization process was determined by myself. The reason why character-level was applied there might be the convenience to compare the translation and annotation results with those from translated English or modern Chinese. However, one single Chinese character can mean more than several words and can be viewed as a highly encoded semantic unit that carries several semantic segments in downstream. Thus, it might be helpful if a more advanced and detailed tokenization is applied in this task, the results can be more heuristic. However, after exploration, at least to my knowledge, there is no related study in decomposition of Chinese characters yet. So, character-level tokenization has to be applied in this task.

Before showing experiments results, the frame and lexical unit names in CFN are translated to English for further comparisons. There are only 12,157 lexical units available in CFN, so several words are added into certain frames from my understanding in the form of *frame.[word.pos_info]*. For example, in [1](a), world '*face*' does not exist in the lexical units of frame 'Posture', but its meaning is actually demonstrated in the frame description. So the first layer should be '*Posture.[face.v]*'. All results are shown in tables below. See Appendix.

For [1](b), the verb '*face*' in Chinese, in addition to meaning of '*with my head facing'*, also conveys the location of '*in front of*'. So here, the last part in CFN should be the '**Depictive**' because it shows the status of the '**Agent**'s' states throughout the time they are in a particular posture. However, in English, the posture '*face*' can only be translated into '*stand in front of*'. So the part '*in front of the pavilion*' is showing the actual location of the posture '*stand*', and this part is annotated as '**Location**'.

For [3](a), the verbs '*hold*', '*stare*' and '*sob*' happen at the same time and at the same level, so they are annotated separately as '**Manipulation.hold.v**','**Perception_active.stare.v**' and '**Make_noise.[sob*.v]**'. For the LU '**hold.v**', parts like '**Agent**', '**Entity**' and '**Bodypart_of_agent**' need to be clarified. '**Entity**' means the **Entity** being manipulated; '**Bodypart_of_agent**' means the part of **Agent**'s body being used to manipulate the **Entity**. However, this sentence can be translated as '*We held each others' hands*', and in this case, '*each others' hands*' can be '**Entity**' and '**Bodypart_of_agent**' at the same time, which can introduce ambiguity. And for the LU '**stare.v**', the phrase '*对视*' can be translated as '*staring into each others' eyes*' and '*对*' here introduces the semantics similar to '*each other*' and '*mutually*' and '*视*' here represents the verb '*stare*' or '*look*'. And at the same time, according to CFN, **Phenomenon** indicates the entity or phenomenon to which the **Perceiver_agentive** directs his or her attention in order to have a perceptual experience. So ‘*对*’ here can be interpreted as **Phenomenon**, and the **Perceiver_agentive** is *DNI*, which can be ‘*we*’ from the context. And for the verb ‘*sob*’ in Chinese, it actually means ‘*cry without noise and choke*’, so itself can introduce the **Degree** of **sob.v** as well. Thus, this case indicates **Degree** can sometimes be included in the verb itself. However, in English translation, the verbs ‘*stare*’ and ‘*choke*’ are used for the additional descriptions of the verb ‘*hold*’. In this case, the all rest part from ‘*teary*’ can be annotate as **Manner** in **Manipulation.hold.v** instead.

For [8](a), in the Chinese sentence, there are two parallel verbs ‘*separate*’ and ‘*ponder*’. And they can be annotated as ‘**Becoming_separated.[separated.v]**’ and ‘**Cogitation.[ponder.v]**’, respectively. For ‘**separate.v**’, ‘*这一*’ can be interpreted as ‘*Once this time*’ in English, which should be annotated as ‘**Time**’ and ‘*常年相别*’ can be understood as ‘*years of separation*’. Thus this part is labeled as ‘**Result**’. For ‘**ponder.v**’, ‘**Cognizer**’ and ‘**Topic**’ need to be clarified, and in this case, ‘**Cognizer**’ is ‘*我(I)*’ and ‘**Topic**’ is ‘*即使遇到好天气、好风景 (Encountering good scenery and good weather though)*’. And ‘**Result**’ is mentioned in the sentence as well, which is ‘*也形同虚设 (will be like nothing as well)*’. While in the English translation, the only frame should be applied is ‘**Categorization**’ and ‘*render*’ is annotated as ‘**Categorization.render.v**’. In this frame, we need to specify the **Cognizer**, the person performs an act of categorization, the **Item**, the entity which is construed or treated by the **Cognizer** as being an instance of a particular **Category**, and the **Criteria**, the general dimensions along which **Items** can potentially differ from one another and hence, fall into different **Category**s. Thus, **Cognizer** here is ‘*Our perennial separation*’, the **Item** here is ‘*any welcoming weather or beautiful scenery that I might encounter*’, and the **Criteria** is ‘*a mere illusion*’.

### 3.3 Applying Chinese FrameNet to Poetry

After experimenting Chinese FrameNet and FrameNet with modern Chinese and translated English, the next experiment step is to apply Chinese FrameNet to Chinese Poetry directly. Before analyzing the final results, semantic segments with similar meanings are marked in the same color. For example, in [1](b), here are the tuples in the same colors: (‘*对*’, ‘*面对着*’, ‘*standing in front of*’), (‘*长亭*’, ‘*长亭*’, ‘*the pavilion*’) and (‘*晚*’, ‘*傍晚时分*’, ‘*In the dusk*’).

For [1](b), the annotation was similar to the former one. The verb ‘*face*’ is missing in the initial CFN data set, so I applied square brackets to notify. In ancient Chinese, one single character can carry complex meanings, and this situation is more general when it comes to poetry. Like the character ‘*对*’ here, in modern Chinese, it only represents the direction(*face against*). However, here it carries not only the direction, but also the motion(*stand*). Thus, I treated ‘*对*’ here as a verb instead of a preposition. And since the poetry was written in a narrative fashion, the subjective can be guessed from the context. So the agent of the verb ‘*对*’ should be *DNI*, which is ‘*我(I)*’ in this case.

For [3](a), there a huge sequence of verbs here, which is [‘*执*’(‘*hold*’), ‘*看*’(‘*stare*’), ‘*泪*’(‘*sob*’), ‘*无语*’(‘*being speechless*), ‘*凝噎*’(‘*choked with tears*’)]. So, before annotation, the relations between each verb need to be specified. Here I assumed all of these verbs happened at the same time and was conducted by the same, or at least part of same subjective. However, there can be more ways to interpret the relations from different time and subjective perspectives, like the translated sentences in modern Chinese and English. In addition, here is an interesting phenomenon, which is ‘noun as verb’, in ancient Chinese. In the final results, the character ‘*泪*’ should be treated as a noun(‘*tears*’ in English) if there is not context. However, in this case, it is a verb which means ‘*使…流泪*’(‘*make something in tears*’ in English). Since this verb can mean make something moisture with tears, so here **Being_wet.[make ... in tears].v** is used to annotate the character ‘*泪*’. Under this frame, an **Item** and a **Liquid** is needed as core FEs: The **Item** takes up the **Liquid** through its pores; the **Liquid** is a fluid or gas that permeates the **Item**. In our case, the **Liquid** FE, which should be ‘tears’, has already been jointly expressed in the verb ‘make ... in tears’.

For [8](a), three sentences show different internal relations between semantic segments. In translated English sentence, in order to keep this sentence more consistent and easy-to-understand, only one subjective is assigned. However, in modern and ancient Chinese sentences, there are more than one subjective involved to create a '*artistic conception*', which I will talk about it in § 4.1. In modern Chinese sentence, a verb 'ponder' is introduced to make the translation more coherent because '*虚设*'(*a mere illusion*) here are considered to be an adjective for '*良辰美景*'(*any welcoming weather or beautiful scenery*). However, this verb does not appear in the initial ancient Chinese sentence. In initial sentence, again '*虚设*'(*a mere illusion*) can be treated as a verb, which should mean '*fake*'. So here **Feigning.[fake.v]** was picked to better represent the verb. And the subjective is '*良辰美景*'(*any welcoming weather or beautiful scenery*).

# 4. Analysis and Conclusions

## 4.1 Features in different genres of Poetry

In this section, I would like to apply CFN to some typical sentences in *Tangshi* and *Yuanqu* for parallel comparisons. Before comparison, here are some important aspects/features that need to be clarified:

a) ***Artistic conceptions***: It is abstract or impressionistic saying of a '*scene*' the author would like to construct poetically. Usually, it consists of several items/subjective with its own motions in parallel time space.

b) ***Syntactic restrains***: As mentioned in § 1, there are different syntactic restrains for different genres of ancient Chinese poetry. For instance, *Tangshi* generally requires each sentence to have a fixed length(5 or 7 characters per sentence); *Songci* and *Yuanqu* requires different lengths of sentences according to its *type of tune*, which means if two pieces of poetry share the same *type of tune*, each of their sentences have the exact same length.

c) ***Phonetic restrains***: No matter in which genres of Chinese poetry, rhythm is one of the most important things that need to be considered seriously. There are five general tunes for vowels in Chinese Pin-yin: for example, ā, á, ǎ, à, a. Although ancient Chinese could pronounce Chinese characters differently from how modern Chinese do, they still share something in common. *Level* (*平*) tones(the first and second tunes) and *oblique* (*仄*) tones(the third and forth tunes) are two patterns for tunes. While composing poetry, the occurrence of both *level* and *oblique* tunes need to be considered according to things like the actual genre of poetry and syntactic restrains. One of the most famous combinations is: '*仄仄平平仄，平平仄仄平。*' In Songci and Yuanqu, sometimes the tunes are fixed because they were served as *lyrics*. And their *type of tunes* were referring one specific type of song that the author was trying to fit.

d) ***Morphological flexibility***: The composition of words in poetry can be invalid if they are used in daily conservation or writing. However, in order to follow the syntactic restrains and phonetic restrains, some concurrences of invalid combinations of characters are allowed to happen.

e) ***Grammatical flexibility***: In ancient Chinese, some nouns and adjectives are used as verbs(see § 3.3), which would bring more flexibility while interpreting sentences. Therefore, the ancient Chinese Poetry offer more possibilities and choices while applying FrameNet. In addition, the order of sentences can be broken in order to better meet both syntactic and phonetic restrains. For example, even post-predicates, which were not considered as legal use of Chinese in general, are allowed to occur as well.

## 4.2 CFN in different genres of Poetry

Two sentences(one for *Tangshi*, one for *Yuanqu*) in two genres of Chinese Poetry will be presented to better illustrate the features/aspects mentioned above.

[1] 国破山河在， 城春草木深。

[2] 枯藤老树昏鸦， 小桥流水人家。

Here are corresponding direct character-level translations:

[1] Country broke homeland exists, city spring grass deep;

[2] Withered vines, old trees, sleepy crows, sophisticated bridges, running creeks, houses.

As we may find in [1], the second sub

sentence is not grammatically legal neither in English nor Chinese. However, it does not mean the author made a mistake in grammar. This phenomenon happened because of *Syntactic restrains* and *Phonetic restrains*. In order to make sense and covey meanings, the sentence leverage the *Grammatical flexibility* and *Morphological flexibility*. First, character '春' initially means the noun '*spring*', however, the author made up another term of use, which is '*something in spring*' as an adverbial. So here the sentence should mean 'c*ity (is) in spring*'. In addition, character '深' means the adjective '*deep*'. However, in this case, this word should be interpreted as a verb, which is '*become deep(er)*'.

[2] should even not be considered as a complete sentence grammatically, since there is no verb involved. And it also is very hard to add one. The author choose to write his poetry like this is due to the concerns of *Phonetic restrains* and *Syntactic restrains.* However, we are still able to understand what the author tried to convey, because he was now constructing an *Artistic conception*. In this *Artistic conception*, items/subjective like *vines*, *trees*, *crows*, *bridges*, *creeks* and *houses* with their own motions, which can be *standing*, *growing* or *running*.

### 4.3 Conclusion

By reviewing the annotation results, we can tell the differences between applications of FrameNet to different languages. And here are some findings I concluded:

1. Chinese verbs can carry its **Degree** with itself sometimes, like '*sob**' in [3](a), which might cause some confusion while applying CFN.

2. In lots of sentences, there is no clear subjective, so that we need to guess from the context. So it brings a lot of *DNI*'s.

3. In some sentences, the multitude of verbs can introduce too much sequential information. The relations between each verb, e.g. whether they are parallel or some of them are used to describe others, like [3](a), can be ambiguous.

4. There are too many different approaches to representing the same semantic meanings using modern Chinese from Ancient Chinese. As a result, CFN annotations are less consistent and stable. It also requires a larger dataset for full annotations than the current CFN data set.

5. One single Chinese character can carry much more semantics than one English word. Thereby, better ways of Chinese decomposition and tokenization are needed for further systematic analysis.

6. Different *restrains* and *flexibilities* make it possible for different interpretation of the same sentences, hence more possible ways to apply FrameNet.

7. The usage of Chinese differs a lot throughout history. So we might need another more specific data set for Ancient Chinese corpus.

## 5. Limitations and Future Work

This work clarifies the application of CFN, and puts forward the findings and patterns found while applying CFN in Ancient Chinese Poetry. Meanwhile, it also gives some insights while conducting multi-lingual FrameNet experiments with translations. However, here exist some limitations as well:

a) Assumption of the good performance of current character-level tokenization for Chinese corpus.
b) Lack of confidence in translation process, the average score of accepted sentence is only 7.786 out of 10.
c) Lack of consideration of all possible interpretations for FrameNet annotations.
d) Low cover rate(50%, see table 2, Appendix) of lexicon units in Chinese FrameNet dataset.

Beyond this work, the further exploration will be focused on guiding the construction of FrameNet task-specific to Chinese Poetry or even Ancient Chinese based on CFN.

## 6. Acknowledgment

I would especially thank Yuansheng Xie, Haotian Xue and Ziyao Ding for their participation in the translation experiments.

# Appendix

1. Translation Results

| sentences | Zirong Chen | Yuansheng Xie | Haotian Xue | Ziyao Ding | Google Translator |
|---|---|---|---|---|---|
| [1](a) | 6.2 | 8.4 | 6.1 | 7.4 | 3.3 |
| [1](b) | 7.8 | 7.4 | 7.6 | 7.6 | 7.3 |
| [1](c) | 7.0 | 7.1 | 7.3 | 7.2 | 7.2 |
| [2](a) | 5.3 | 8.2 | 6.0 | 6.2 | 4.6 |
| [2](b) | 7.4 | 7.4 | 6.0 | 7.6 | 4.3 |
| [3](a) | 5.1 | 7.0 | 6.7 | 6.4 | 4.1 |
| [4](a) | 6.4 | 8.1 | 6.1 | 6.7 | 3.2 |
| [4](b) | 6.5 | 8.6 | 6.6 | 6.1 | 5.5 |
| [5](a) | 7.8 | 8.1 | 7.8 | 7.6 | 5.3 |
| [5](b) | 7.4 | 7.1 | 7.0 | 7.2 | 6.2 |
| [6](a) | 7.4 | 7.9 | 8.1 | 8.2 | 6.7 |
| [7](a) | 7.4 | 7.6 | 7.8 | 7.1 | 4.2 |
| [8](a) | 6.4 | 7.1 | 6.0 | 6.6 | 2.9 |
| [9](a) | 7.4 | 7.1 | 7.0 | 7.3 | 3.7 |
| Avg Score | 6.83 | 7.65 | 6.86 | 7.09 | 4.90 |

Table 1

2. FrameNet Annotations

[1](b)

对长亭晚

| | NI | 对 | 长亭 | 晚 |
|---|---|---|---|---|
| Posture.[face.v] | | face | | |
| FE | Agent(DNI) | | Depicitive | Time |

傍晚时分，面对着长亭。

| | NI | 傍晚 | 时分 | ， | 面 | 对着长亭 |
|---|---|---|---|---|---|---|
| Posture.[face.v] | | | | | face | |
| FE | Agent(DNI) | Time | | | | Depicitive |

In the dusk, (I was) standing in front of the pavilion.

| | NI | In | the | dusk | , | standing | in | front | of | the | pavilion |
|---|---|---|---|---|---|---|---|---|---|---|---|
| Posture.stand.v | | | | | | stand | | | | | |
| FE | Agent(DNI) | Time | | | | | Location | | | | |

[3](a)

执手相看泪眼，竟无语凝噎。

| | NI | 执 | 手 | 相 | 看 | 泪 | 眼 | ， | 竟 | 无语 | 凝噎 |
|---|---|---|---|---|---|---|---|---|---|---|---|
| Manipulation.hold.v | | hold | | | | | | | | | |
| FE | Agent(DNI) | | Entity | | | | | | | | |
| Perception_active.stare.v | | | | | stare | | | | | | |
| FE | Perceiver_agentive(DNI) | | | Phenomenon | | | | | | | |
| Being_wet.[make...in tear].v | | | | | | Make...in tear | | | | | |
| FE | | | | | | Liquid* | Item | | | | |
| Volubility.silent.a | | | | | | | | | | silent | |
| FE | Speaker(DNI) | | | | | | | | | | |
| Make_noise.sob.v | | | | | | | | | | | sob |
| FE | Sound_source(DNI) | | | | | | | | | | Degree* |

握着对方的手含着泪对视，哽咽的说不出话来。

| | NI | 握着 | 对方的手 | 含着泪 | 对 | 视 | ， | 哽咽 | 得说不出话来 |
|---|---|---|---|---|---|---|---|---|---|
| Manipulation.hold.v | | hold | | | | | | | |
| FE | Agent(DNI) | | Entity | | | | | | |
| Perception_active.stare.v | | | | | | stare | | | |
| FE | Perceiver_agentive(DNI) | | | Depictive | Phenomenon | | | | |
| Make_noise.sob.v | | | | | | | | sob | |
| FE | Sound_source(DNI) | | | | | | | Degree* | Sound |

We held each others' hands, teary eyes staring, choked with sobs and speechless.

| | NI | We | held | each | others' | hands | , | teary | eyes | staring | , | choked | with | sobs | and | speechless |
|---|---|---|---|---|---|---|---|---|---|---|---|---|---|---|---|---|
| Manipulation.hold.v | | | hold | | | | | | | | | | | | | |
| FE | | Agent(DNI) | | Entity | | | | Manner | | | | | | | | |

[8](a)

此去经年，应是良辰好景虚设

| | NI | 此 | 去 | 经 | 年 | ， | 应是 | 良辰 | 美景 | 虚设 |
|---|---|---|---|---|---|---|---|---|---|---|
| Becoming_separated.[separate.v] | | | separate | | | | | | | |
| FE | Whole(DNI) | Time | | Result | | | | | | |
| Feigning.[fake.v] | | | | | | | | | | fake |
| FE | | | | | | | | Agent | | |

这一去长年相别，我料想即使遇到好天气、好风景，也如同虚设

| | NI | 这一 | 去 | 常年 | 相别 | ， | 我 | 料想 | 即使 | 遇到的 | 好天气 | 、 | 好风景 | ， | 也 | 如同 | 虚设 |
|---|---|---|---|---|---|---|---|---|---|---|---|---|---|---|---|---|---|
| Becoming_separated.[separate.v] | | | separate | | | | | | | | | | | | | | |
| FE | Whole(DNI) | Time | | Result | | | | | | | | | | | | | |
| Cogitation.[ponder.v] | | | | | | | | ponder | | | | | | | | | |
| FE | | | | | | | Cognizer | | Topic | | | | | | Result | | |

Our perennial separation renders any welcoming weather or beautiful scenery that I might encounter a mere illusion.

| | NI | Our | perennial | separation | renders | any | welcoming | weather | or | beautiful | scenery | that | I | might | encounter | a | mere | illusion |
|---|---|---|---|---|---|---|---|---|---|---|---|---|---|---|---|---|---|---|
| Categorization.render.v | | | | | render | | | | | | | | | | | | | |
| FE | | Cognizer | | | | Item | | | | | | | | | | Criteria | | |

3. Chinese FrameNet Cover Rate

| | All FE layers | Missing LUs | Cover Rate |
|---|---|---|---|
| Ancient Chinese | 8 | 4 | 50% |
| Modern Chinese | 6 | 3 | 50% |

Table 2